\documentclass[]{interact}

\usepackage{epstopdf}
\usepackage{subfigure}

\usepackage{amsmath,amssymb,amsfonts}
\usepackage{algorithmic}
\usepackage{graphicx}
\usepackage{textcomp}

\usepackage[utf8]{inputenc} 
\usepackage[T1]{fontenc}    
\usepackage{url}            
\usepackage{booktabs}       
\usepackage{nicefrac}       
\usepackage{microtype}      
\usepackage{xcolor}         
\usepackage{blindtext}
\usepackage{multirow}
\usepackage{lipsum}
\usepackage{amsthm}
\usepackage{tgpagella}
\usepackage{emptypage}
\usepackage[LGR,T1]{fontenc}
\usepackage{hyperref}

\usepackage{natbib}
\bibpunct[, ]{(}{)}{;}{a}{}{,}
\renewcommand\bibfont{\fontsize{10}{12}\selectfont}

\theoremstyle{plain}

\theoremstyle{definition}

\theoremstyle{remark}

\usepackage{amsmath,amsfonts,bm}

\def\eqref#1{equation~\ref{#1}}

\def\1{\bm{1}}

\DeclareMathAlphabet{\mathsfit}{\encodingdefault}{\sfdefault}{m}{sl}
\SetMathAlphabet{\mathsfit}{bold}{\encodingdefault}{\sfdefault}{bx}{n}

\def\gE{{\mathcal{E}}}

\def\gG{{\mathcal{G}}}

\def\gV{{\mathcal{V}}}

\begin{document}

\articletype{RESEARCH ARTICLE}

\title{Graph Neural Networks for Supply Chain Management: A Framework and Real World Benchmark}

\author{
\name{Azmine Toushik Wasi\textsuperscript{a}\thanks{CONTACT A.~T. Wasi. Email: azmine32@student.sust.edu} and
MD Shafikul Islam\textsuperscript{b}} 
\affil{
\textsuperscript{a}Department of Industrial and Production Engineering, Shahjalal University of Science and Technology, Sylhet, Bangladesh; \\
\textsuperscript{b}Department of Mechanincal and Industrial Engineering, Louisiana State University, Baton Rounge, LA, 70803, USA;
}
}

\maketitle

\begin{abstract}
Graph Neural Networks (GNNs) have recently gained traction in transportation, bioinformatics, language and image processing, but research on their application to supply chain management remains limited. Supply chains are inherently graph-like, making them ideal for GNN methodologies, which can optimize and solve complex problems. 
The barriers include a lack of proper conceptual foundations, familiarity with graph applications in SCM, and real-world benchmark datasets for GNN-based supply chain research. To address this, we discuss and connect supply chains with graph structures for effective GNN application, providing detailed formulations, examples, mathematical definitions, and task guidelines. Additionally, we present a multi-perspective real-world benchmark dataset from a leading FMCG company in Bangladesh, focusing on supply chain planning.
We discuss various supply chain tasks using GNNs and benchmark several state-of-the-art models on homogeneous and heterogeneous graphs across six supply chain analytics tasks. Our analysis shows that GNN-based models consistently outperform statistical Machine Learning and other Deep Learning models by around 10-30\% in regression, 10-30\% in classification and detection tasks, and 15-40\% in anomaly detection tasks on designated metrics.
With this work, we lay the groundwork for solving supply chain problems using GNNs, supported by conceptual discussions, methodological insights, and a comprehensive dataset.
\end{abstract}


\section{Introduction}
Supply chain is a dynamic network of organizations that participate in the various processes and activities that produce value in the form of products and services for consumers via upstream and downstream linkages \citep{mentzer2001defining}, entailing a continuous flow of information, goods, and money among its various stages \citep{higuchi2004dynamic}. As the supply chain consists of interconnected entities in a complex network, it involves intricate interdependencies and complex decision-making processes \citep{surana2005supply}. Additionally, the modern supply chain generates enormous data, with relationships and dependencies between entities requiring sophisticated models to capture \citep{sadeghiamirshahidi2014improving}. The potential benefits of using computational methods to solve supply chain problems include improved coordination, efficient logistics, and effective supply chain solutions \citep{chaovalitwongse2013computational}.

\begin{figure} 
\centering {\includegraphics[width=\textwidth]{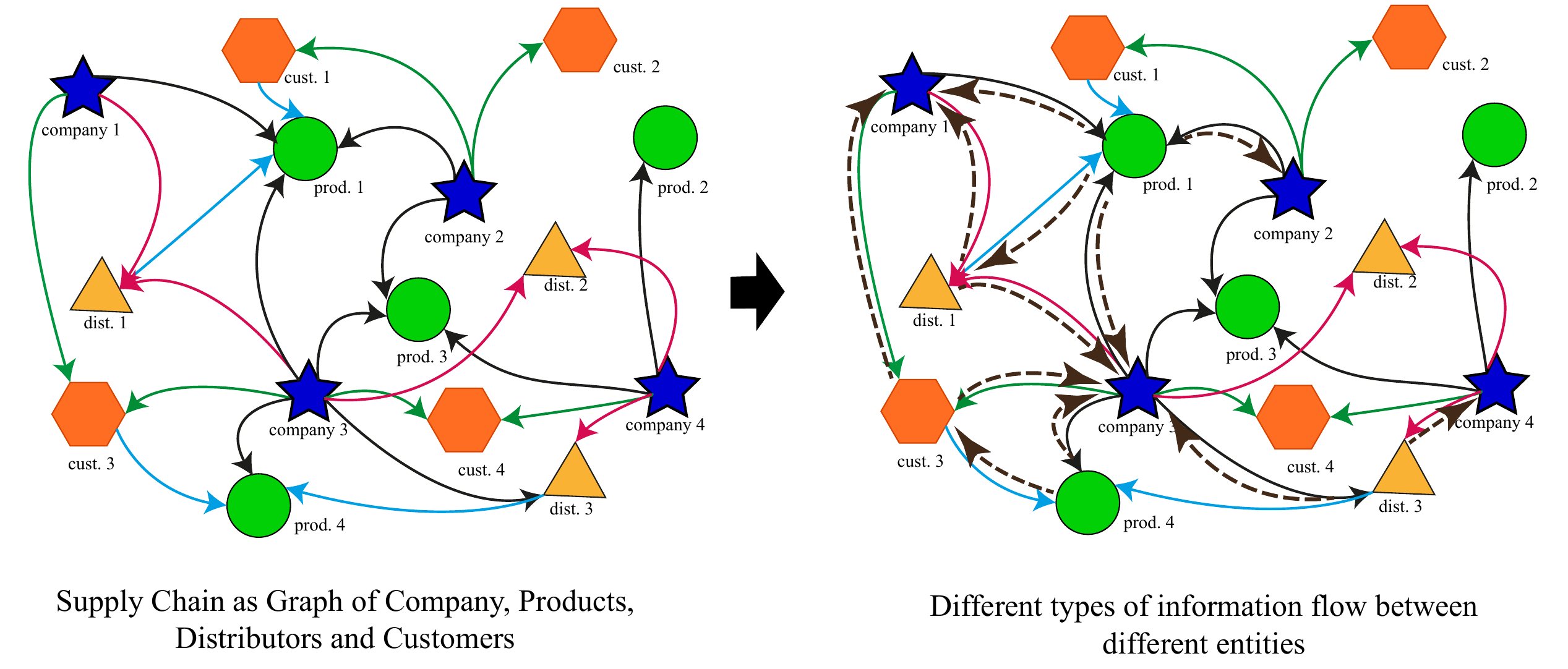}}
\caption{Supply Chain as a Graph of Interconnected Company, Products, Distributors and Customers.}\label{fig:what-m}
\end{figure}

Graph Neural Networks (GNNs) are widely used across various sectors due to their ability to model complex relationships in data. In social networks, GNNs help analyze user connections for recommendations, information analysis and community detection \cite{wasi-etal-2024-banglaautokg}. In biology, they assist in drug discovery by modelling molecular interactions \citep{Johnson2024, wasi2024cadglcontextawaredeepgraph}. 
GNNs are used in finance for fraud detection \citep{wu2024heterogeneous}, in HR for job matching \cite{wasi-2024-hrgraph}, and in research for knowledge graph reasoning \cite{kosasih2022towards} and anomaly detection \cite{8276segewagt899,zhou2023hktgnn}. Their strength lies in effectively capturing dependencies in non-Euclidean data, making them versatile for tasks involving connected entities.
Connecting to supply chain, GNNs can enable the modelling of complex relationships and dependencies within the supply chain, facilitating tasks such as sales predictions, production planning, risk assessment, and uncovering hidden risks \citep{zhou2023hktgnn, kosasih2022towards}. By leveraging GNN methodologies, it is possible to optimize supply chain operations, enhance risk management, and improve decision-making processes by extracting relevant information from graph data and inferring multiple types of hidden relationship risks \citep{kosasih2022towards}.

Production planning plays a pivotal role in supply chain management by forecasting future product or service demand, aiding organizations in optimizing inventory levels, production schedules, and resource allocation \citep{aamer2020data, zougagh2020prediction}. The accuracy of demand prediction significantly impacts company revenue, prompting the exploration of diverse deep learning and machine learning models \citep{alves2021applying, pirhooshyaran2020simultaneous, pacella2021evaluation}. 
While traditional models have shown promise, GNNs offer a unique advantage in modelling the network-like structures inherent in supply chains, such as global trade flows or social networks \citep{kosasih2022machine}, as shown in Figure \ref{fig:what-m}. Despite limited prior studies on GNNs in supply chains, recent research has demonstrated their utility in tasks like hidden link prediction to mitigate risk and uncover hidden dependencies \citep{aziz2021data}. However, several challenges remain: there is a lack of comprehensive conceptual foundations and formulations specific to supply chain applications of GNNs, and researchers often lack awareness of the diverse tasks that GNNs can address in this domain. Moreover, the scarcity of publicly available datasets and proper benchmarks hinders thorough evaluation and development of GNN models for supply chains. These gaps underscore the need for detailed methodologies and robust datasets to advance research and practical applications of GNNs in supply chain optimization.


To address these challenges, we take several key steps. First, we thoroughly discuss and connect supply chains with graph structures, formulating them as graphs for effective GNN application. We provide detailed definitions and explanations of how supply chain components map to graph properties, illustrating how GNNs can utilize these properties to tackle supply chain problems efficiently. Second, we introduce a new, multi-perspective benchmark dataset tailored for GNN applications in supply chain planning, accompanied by a comprehensive exploratory study of the dataset. Third, we explore a variety of supply chain tasks from different graph perspectives and benchmark state-of-the-art models on both homogeneous and heterogeneous graphs across six distinct analytics and modelling tasks. Our experiments demonstrate that GNN-based models significantly outperform traditional statistical and machine learning methods in various supply chain applications, underscoring the vital role of GNNs in advancing supply chain analysis and modelling.

\vspace{6mm}
The contributions of this research are summarized into these key aspects:
\begin{itemize}
\item We comprehensively discuss and connect supply chains with graphs, and formulate supply chains as graphs for effective application of GNNs. We define and explain how various supply chain components correspond to graph properties, and demonstrate how GNNs can leverage these properties to solve supply chain problems efficiently.
\item We introduce a new, multi-perspective benchmark dataset for GNN problem-solving in supply chain planning, paving the way for extensive research and analytics using GNNs in supply chain management. Additionally, we present a comprehensive exploratory study of the dataset. The dataset is publicly available at \textbf{\textit{\href{https://doi.org/10.5281/zenodo.13652826}{\texttt{DOI: 10.5281/zenodo.13652826}}}} under the \href{https://creativecommons.org/licenses/by/4.0/legalcode}{CC BY 4.0} Licence.
\item We describe and explore various tasks of supply chain from different graph perspectives like homogeneous graph, heterogeneous graph and hypergrphs; bench-marking the performance of state-of-the-art models on both homogeneous and heterogeneous graphs across six different supply chain analytics and modelling tasks on the dataset.
\item Our dataset analysis and modelling experiments show that graph-based models outperform traditional statistical and machine learning methods in various supply chain applications, highlighting the critical role of GNNs in effective supply chain analysis and modelling.
\item We also discuss real-life implications of the work, addressing their societal impacts and offering insights into potential future research directions; along with limitations of the current study and supply chain as graph methods.
\end{itemize}

\vspace{6mm}
With these contributions, we aid to lay the groundwork for advancing the application of GNNs in supply chain contexts. The following sections will review the related works and provide a detailed analysis of machine learning and GNN approaches in supply chain management.

\section{Related Works} 
Supply chain management has seen significant advancements through the application of machine learning and graph neural networks. This section reviews the existing literature on machine learning approaches and the emerging role of GNNs in supply chain management, identifying research gaps and potential areas for future exploration. 

\subsection{Machine Learning (ML) Approaches in Supply Chain}

ML models aid in demand forecasting, inventory management, route optimization, supplier risk assessment, and quality control, leading to improved operational efficiency and cost reduction \citep{polo2024integration}. By applying AI algorithms to the supply chain management system, organizations can achieve visualization, automation, and intelligent management of all supply chain links, ultimately improving responsiveness to market demands and reducing operating costs \citep{lin2022innovative}. Extensive research has been dedicated to harnessing ML \citep{feizabadi2022machine, zhu2021demand, aamer2020data}. Numerous studies aim to improve demand prediction and optimize production processes through these methodologies \citep{nitsche2021mapping, younis2022applications, filali2021exploring}.

Deep learning techniques, including ANNs and CNNs, have been widely explored in supply chains. ANNs are effective in modelling non-linear relationships and adapting to historical data, though their performance is influenced by parameters like neuron count and learning rate \citep{lunardi2021comparison, mrad2019improved}. While Random Forest classifiers often outperform ANNs in demand forecasting, ANNs remain close in accuracy \citep{vairagade2019demand}.
CNNs have been used for various supply chain tasks, with studies showing LSTM models generally outperform CNNs for grocery sales forecasting \citep{husna2021demand}. CNNs are useful for improving inventory forecasts and reducing costs by enhancing supplier preparation \citep{tang2022cnn}. Comparative analyses indicate that CNNs slightly outperform LSTMs in some datasets \citep{bousqaoui2021comparative}. Extreme gradient boosting models also achieve high precision with enhanced features \citep{lingelbach2021demand}.

While no single model consistently demonstrates superior performance across all scenarios, various models excel in different situations based on the problem type and specific context. This underscores the potential of GNNs to provide robust solutions in scenarios where traditional models show variable performance.

\subsection{Graph Neural Networks in Supply Chain}

Recent advancements in GNNs have enhanced demand forecasting and operational resilience, leading to more efficient and adaptive supply chain management. Graph representation learning has further improved link prediction by uncovering hidden dependencies within supply networks \citep{aziz2021data, kosasih2022machine} and building on previous GNN link prediction research \citep{zhang2018link, Zhang2017WeisfeilerLehmanNM, Cai2020LineGN, Cai2020AMA}. These developments highlight the effectiveness of machine learning in refining demand forecasting and production planning. Notable contributions include GNN-based hidden link prediction for risk mitigation and integrating GNNs with knowledge graph reasoning to identify latent risks and extract insights \citep{aziz2021data, kosasih2022machine, kosasih2022towards}.

GNNs have been applied for supplier recommendations by analysing network data to suggest alternative suppliers during disruptions \citep{tu2024using}. The Hierarchical Knowledge Transferable Graph Neural Network (HKTGNN) simplifies complex supply chains and uses a centrality-based knowledge transfer module for risk assessment \citep{zhou2023hktgnn}. GNNs have also enhanced the accuracy of industry classification based on supply chain data \citep{wu2023industry} and been used in heterogeneous graph neural networks for fraud detection and explanation in supply chain finance, leveraging multi-view information \citep{wu2024heterogeneous}.

\subsection{Research Gap Analysis}
While the potential of machine learning in supply chain management is widely acknowledged, the research focus has only recently shifted to Graph Neural Networks (GNNs) due to their unique capabilities in modelling supply chain networks \citep{song2021network}. However, the application of GNNs within supply chain research remains relatively nascent. \citet{916492aegfg3} undertook an analysis of supply chains from a graph-theoretical standpoint, employing measures such as node centrality, clustering, and other statistical graph approaches. Notably absent from their study were any methods based on GNNs. Similarly, Wagner and Neshat \citep{WAGNER2010121} pursued a comparable analysis using graph-theoretic techniques, yet they also did not incorporate GNN-based methodologies in their work.

Despite the growing recognition of GNNs' potential, progress in GNN-based supply chain research has been slow. This highlights the significance of specialized datasets and cutting-edge methodologies for fully utilizing GNNs' advantages in addressing complex demand prediction and production planning issues in the supply chain domain.
Various supply chain problems have been addressed using GNNs, such as hidden link prediction tasks \citep{aziz2021data, kosasih2022machine}, supplier recommendation \citep{tu2024using}, risk mitigation \citep{kosasih2022towards}, supply chain risk assessment \citep{zhou2023hktgnn}, industry classification \citep{wu2023industry}, and fraud detection and explanation in supply chain finance \citep{wu2024heterogeneous}. However, production planning, one of the most critical tasks in supply chain management that ensures profitability and is influenced by numerous internal and external factors, has not yet been effectively addressed using GNNs. This represents a significant untapped potential for GNN-based methodologies to provide superior solutions in this domain.

To address this gap, we propose SCG, a benchmark dataset designed specifically for GNN-based supply chain analysis. This dataset aims to explore the potential of GNN models in production planning, demonstrating their capability to handle complex supply chain structures and improve decision-making processes. The introduction of this dataset is intended to spur further research and development in this promising area, leveraging GNNs to achieve more efficient and adaptive supply chain operations.

\vspace{6mm}

Following the review of related works, the terminology and foundational concepts of graph theory are introduced. In the section \ref{sec:Sc-aG}, titled \textit{Supply Chain Problem Formulation in Graphs}, the supply chain problem is rigorously defined and formulated within a graph-based framework. Subsequently, the \textit{Data Collection} section details the process of gathering and organizing data from a leading FMCG company in Bangladesh, ensuring a robust dataset for analysis. This is followed by a detailed description and statistical analysis of the dataset. The \textit{Dataset Potential and Practical Applications} section explores the dataset's versatility in addressing various supply chain challenges. Experimental results demonstrating the effectiveness of GNNs in this domain are showcased in the \textit{Experiments} section. The study concludes with a discussion of the findings, their implications, and future research directions and possibilities.

\section {Graph Neural Networks}
Graphs can be categorized into different types, including homogeneous graphs, where all nodes and edges are of the same type, and heterogeneous graphs, which involve multiple types of nodes and edges. Each type is tailored for specific applications, depending on the structure and relationships represented. Homogeneous graphs are ideal for simpler relationships, such as a social network where nodes represent people and edges represent friendships. Heterogeneous graphs, however, feature multiple types of nodes and edges, capturing more complex interactions, like in a business network with nodes representing employees, departments, and projects. Hypergraphs extend this complexity further by allowing hyperedges to connect more than two nodes, useful for modelling multi-party interactions.
Different types of GNNs designed for different types of graphs can be used in supply chain applications.

\vspace{2mm}
Here, we define and discuss graphs and different types of GNNs, along with their pros and cons, in relation to supply chain optimization applications.

\subsection {Definition of Graphs and GNNs} \label{sec:gnn-def}
A {\em graph} $\gG = (\gV, \gE)$ is a collection of {\em nodes} $\gV$  and {\em edges} $\gE \subseteq \gV\times \gV$ between pairs of nodes. For the purpose of the following discussion, we will further assume the nodes to be endowed with $s$-dimensional {\em node features}, denoted by $\mathbf{x}_u$ for all $u \in \gV$. 

We consider a graph to be specified with an adjacency matrix $\mathbf{A}$ and node features $\mathbf{X}$. In the generic setting $\gE\neq\emptyset$, the graph connectivity can be represented by the $n\times n$ {\em adjacency matrix} $\mathbf{A}$, defined as  
\begin{equation}
    a_{uv} = \begin{cases}
    1 & (u, v)\in \gE\\
    0 & \text{otherwise}.
    \end{cases}
\end{equation}
Here, the adjacency and feature matrices $\mathbf{A}$ and $\mathbf{X}$ are synchronized, in the sense that $a_{uv}$ specifies the adjacency information between the nodes described by the $u$th and $v$th rows of $\mathbf{X}$. 

\vspace{4mm}
A \textit{Graph Neural Network (GNN)} is a type of neural network designed to operate on graph structures. In a graph, nodes represent entities, and edges represent relationships between these entities. GNNs leverage the connections and relationships within the graph to learn node representations, enabling tasks like node classification, link prediction, graph classification, and many more. By aggregating information from neighbouring nodes, GNNs can capture both local and global graph structure, making them effective for complex relational data \citep{DBLP:journals/corr/abs-2104-13478,Hamilton2020,asefgawegwgwg}.

\subsection {Classification of GNNs}
As discussed by \citet{DBLP:journals/corr/abs-2104-13478}, GNNs can be described into three main types or, flavours. They are: Convolutional,  Attentional, and Message-passing.

\subsubsection{Convolutional GNNs}
Convolutional GNNs \citep{kipf2017semisupervised,defferrard2016convolutional,wu2019simplifying} aggregate features from neighbouring nodes using fixed weights.
In this class, the features of the neighbourhood nodes are  directly aggregated with fixed weights,
\begin{equation}
    \vec{h}_u = \phi\left(\vec{x}_u, \bigoplus\limits_{v\in\mathcal{N}_u}c_{uv}\psi(\vec{x}_v)\right).
\end{equation}
Here, $c_{uv}$ specifies the \emph{importance} of node $v$ to node $u$'s representation. It is a constant that often directly depends on the entries, and ${\bf A}$ representing the structure of the graph. 

 This method is straightforward and computationally efficient, making it suitable for large-scale graphs. However, it often overlooks the varying importance of different neighbours, potentially leading to suboptimal performance in scenarios where certain relationships are more significant than others. In supply chain analytics, Convolutional GNNs can be used for tasks like demand forecasting and inventory optimization, where the relationships between various entities (e.g., suppliers and distributors) are relatively uniform.

\subsubsection{Attentional GNNs}
Attentional GNNs address the limitations of Convolutional GNNs by introducing learnable self-attention mechanisms that assign varying importance to different neighbours.
In this flavour \citep{Velickovic:2018we,monti2017geometric,zhang2018gaan}, the interactions are implicit: 
\begin{equation}
    \vec{h}_u = \phi\left(\vec{x}_u, \bigoplus\limits_{v\in\mathcal{N}_u}a(\vec{x}_u, \vec{x}_v)\psi(\vec{x}_v)\right).
\end{equation}
Here, $a$ is a learnable \emph{self-attention mechanism}  that computes the importance coefficients $\alpha_{uv} = a(\vec{x}_u, \vec{x}_v)$ implicitly. They are often softmax normalised across all neighbours. 
When $\bigoplus$ is the summation, the aggregation is still a linear combination of the neighbourhood node features, but now the weights are feature-dependent. 

 This approach allows the model to focus on more relevant connections, improving prediction accuracy. The downside is the increased computational cost and complexity. In supply chain analytics, Attentional GNNs are particularly useful for anomaly detection and risk management, where the ability to prioritize critical relationships can lead to more accurate insights.

\subsubsection{Message-passing GNNs}
Message-passing GNNs \citep{gilmer2017neural,battaglia2018relational} take a more flexible approach by allowing arbitrary messages to be passed across edges. It amounts to computing arbitrary vectors \textit{(messages)} across edges, 
\begin{equation}
    \vec{h}_u = \phi\left(\vec{x}_u, \bigoplus\limits_{v\in\mathcal{N}_u}\psi(\vec{x}_u, \vec{x}_v)\right).
\end{equation}
Here, $\psi$ is a learnable \emph{message function}, computing $v$'s vector sent to $u$, and the aggregation can be considered as a form of message passing on the graph.

This method is highly expressive and can capture complex dependencies within the graph. However, it can be computationally intensive and may require substantial training data to achieve good performance. In supply chain analytics, Message-passing GNNs can be applied to production forecasting and dynamic routing, where the ability to model intricate interactions is crucial.

\subsection{Temporal GNNs} 
Another interesting and effective GNN model for supply chain analytics is the Temporal Graph Neural Network (Temporal GNN) \citep{Zhao2018TemporalGC, rossi2020temporalgraphnetworksdeep,  longa2023graphneuralnetworkstemporal}. Temporal GNNs extend traditional GNNs by incorporating the dimension of time, allowing them to model dynamic graphs where relationships and features change over time. A general equation for Temporal GNNs can be expressed as:

\begin{equation} 
\mathbf{h}_v^{(t+1)} = \sigma \left( \sum_{u \in \mathcal{N}(v)} \mathbf{A}_{uv}^{(t)} \mathbf{h}_u^{(t)} \mathbf{W}^{(t)} + \mathbf{B}^{(t)} \mathbf{h}_v^{(t)} \right)
\end{equation}

where \(\mathbf{h}_v^{(t)}\) represents the hidden state of node \(v\) at time \(t\), \(\mathcal{N}(v)\) denotes the neighbors of \(v\), \(\mathbf{A}_{uv}^{(t)}\) is the temporal adjacency matrix, \(\mathbf{W}^{(t)}\) and \(\mathbf{B}^{(t)}\) are learnable weight matrices, and \(\sigma\) is an activation function. Temporal GNNs are crucial for modelling dynamic systems where relationships and features evolve over time. In supply chain analytics, they enable the prediction of time-dependent phenomena such as demand fluctuations, shipment delays, and inventory levels. By capturing temporal patterns, these models provide insights into future trends, helping businesses optimize operations and anticipate disruptions. 

\textbf{Spatio-Temporal GNNs}, a subset of Temporal GNNs, combine spatial and temporal information, making them exceptionally powerful for supply chain analytics. These models can capture not only the evolving relationships between entities but also their spatial dependencies \citep{ zhang2018gaan, rozemberczki2021pytorch}. In supply chains, Spatio-Temporal GNNs can significantly enhance predictive accuracy for tasks such as demand forecasting, dynamic route optimization, and real-time inventory management. Using both spatial and temporal dynamics, these models enable more informed decision-making, leading to improved efficiency and resilience in supply chain operations.

\section{Supply Chain Problem Formulation in Graphs}\label{sec:Sc-aG}
Supply chains consist of numerous interconnected components, including production facilities, products, and raw materials. There are various production facilities in an industry, each responsible for producing different types of products or products of different stock keeping units (SKUs). Each component, whether a different production facility or a different product group within the same production facility, can be represented as a node in a graph. Connections (edges) within this graphical model represent the relationships among the nodes. Products within the same group or produced in the same facility share connections due to common factors such as production capacity, raw material requirements, and demand trends. Additionally, production facilities may be interrelated due to shared resources, production dependencies, or logistical connections. By employing a graph-based approach, these complex relationships can be systematically analysed. This graphical representation allows for the integration of diverse data attributes for each node, including demand, production capacity, and sales metrics, thereby enabling a holistic analysis of the supply chain network.

\begin{figure} 
\centering {\includegraphics[width=\textwidth]{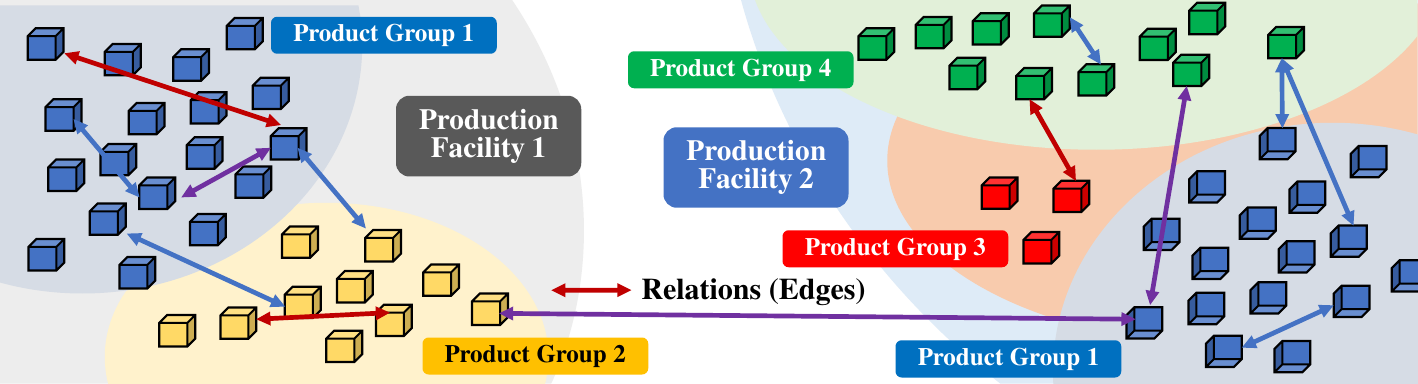}}
\caption{\textbf{Supply Chain Problem Formulation in Homogeneous Graph.} Boxes represent various product types, with colour indicating different groups. They are closely located based on product groups and production facilities. Different relational connections denote shared raw material requirements, interdependence between products, and other impacts.}\label{fig:1-formulation}
\end{figure}

To illustrate the concept, consider a potato chip manufacturing company. This company operates two production facilities and produces four types of chips: normal chips, triangular chips, ring chips, and string chips, each represented by different colours in Figure \ref{fig:1-formulation}. Each type of chip is available in various packaging sizes, such as 30g, 50g, and 100g, depicted as different boxes in the figure. Within this graphical model, connections represent the relationships among the chips, facilities, and production processes. For instance, products within the same group or produced in the same facility share connections due to common factors such as production capacity, raw material requirements, and demand trends. If the production capacity for a chip group is limited to 500kg per day, increasing the production of 500g packets may necessitate reducing the output of 30g packets, demonstrating an internal group dependency. Moreover, if a machine can only produce one flavour at a time, producing multiple flavours on the same day would be inefficient due to the time lost in changeovers. If one facility specializes in a particular chip type, producing different types in the same facility may lead to inefficiencies. These interdependencies are depicted as connections within the same facility in Figure \ref{fig:1-formulation}.Inter-facility relationships also exist, where producing a specific chip type in one facility may impact the production capabilities of another facility. Shared raw materials between product groups can create dependencies, where the production of one group affects the other.

\begin{figure} 
\centering {\includegraphics[width=\textwidth]{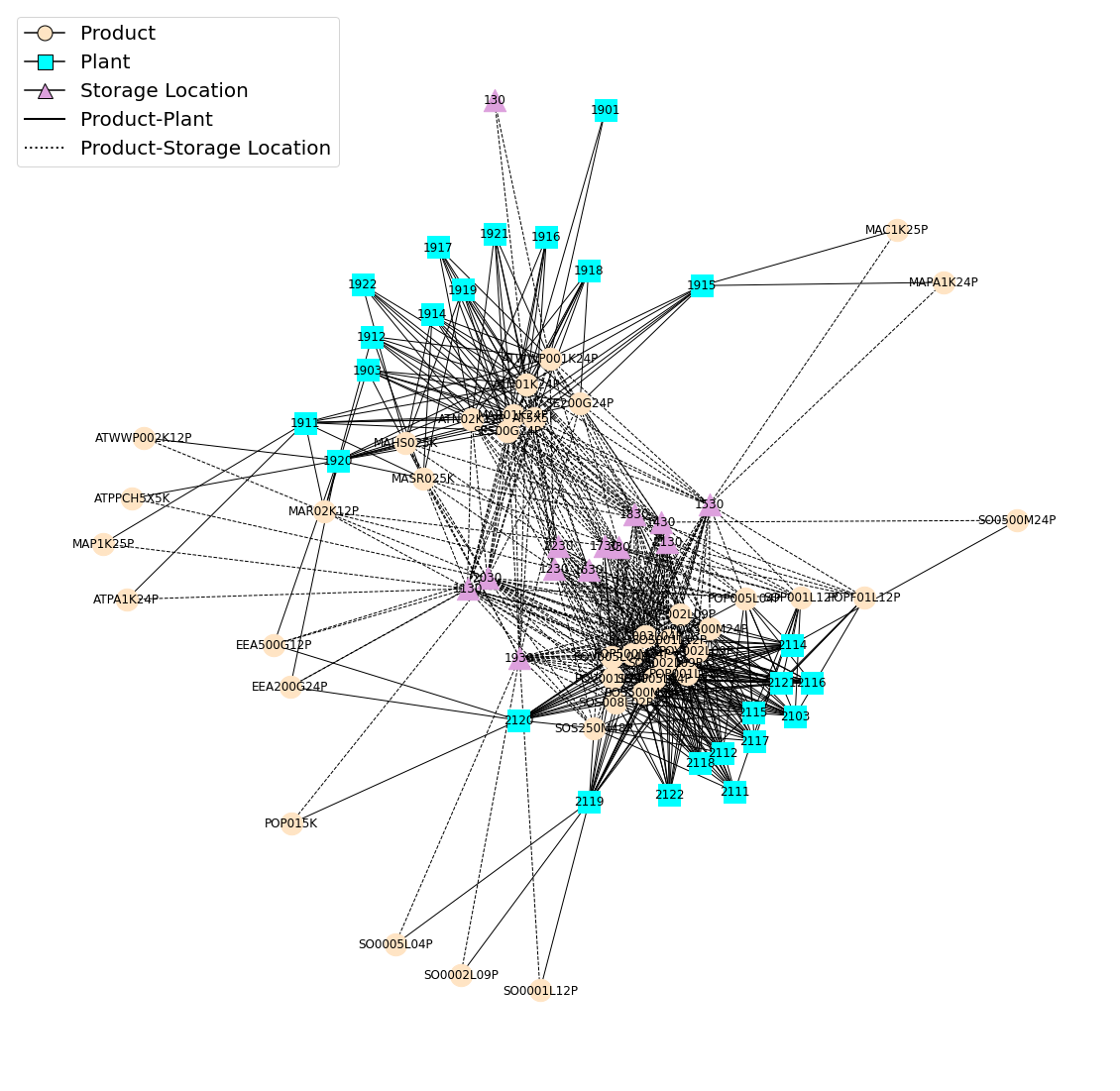}}
\caption{Heterogeneous Graph Example using SCG dataset. Products, plants, and storage locations are nodes and their relations are edges.}\label{fig:heterogeneous-graph}
\end{figure}

By utilizing a graph-based approach, these complex relationships can be systematically analysed. GNNs leverage this structured representation to enhance production forecasting and planning, offering a comprehensive solution to supply chain management challenges.Formulating supply chain problems as graphs enables the visualization and analysis of intricate relationships among supply chain elements. Figure \ref{fig:heterogeneous-graph} illustrates a heterogeneous graph representing the supply chain network. It includes nodes for products (beige), plants (cyan), and storage locations (purple), with edges depicting the relationships between these entities. Solid lines indicate connections between products and plants, while dashed lines represent links between products and storage locations, showcasing the complexity and interdependencies within the supply chain.

\subsection{Defining Supply Chain as Graph}
Following discussions on supply chain as graph in Section \ref{sec:Sc-aG} and general GNN formulations in Section \ref{sec:gnn-def}, 
we can formulate a supply chain graph $\gG = (\gV, \gE)$ by considering each product as a node $u \in \gV$, and the connections between products—such as being manufactured in the same factory or belonging to the same product group or subgroup as edges $(u, v) \in \gE$. The nodes are equipped with $s$-dimensional feature vectors $\mathbf{x}_u$, which capture the production and demand history of each product. This means that for each product (node), we have an associated feature vector $\mathbf{x}_u$ that contains historical data on its production quantities and demand patterns. The graph connectivity is represented by an $n \times n$ adjacency matrix $\mathbf{A}$. $a_{uv}$ from $\mathbf{A}$ specifies the connectivity between the products represented by the $u$Th and $v$Th rows of $\mathbf{X}$. This representation helps in understanding and analysing the relationships and dependencies between different products in the supply chain, providing insights into production efficiency and demand forecasting.

\begin{figure}
\centering
\subfigure[Here, the nodes are sub-group products, and plants are edges. Colors denote different types of nodes and edges.]{%
\resizebox*{7cm}{!}{\includegraphics[scale=0.35]{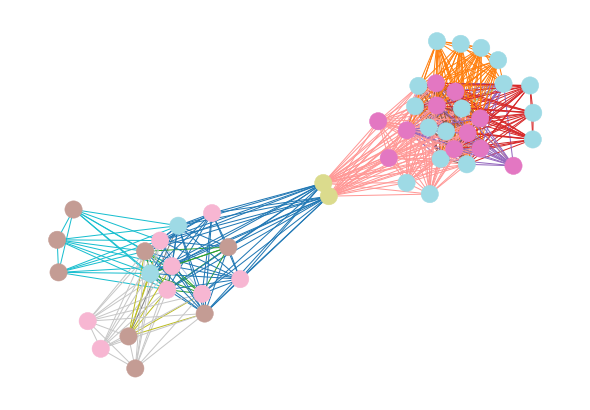}}}\hspace{5pt}
\subfigure[Here, nodes plant-products, and storage locations
are edges. Colors denote different types of nodes and edges.]{%
\resizebox*{7cm}{!}{\includegraphics[scale=0.35]{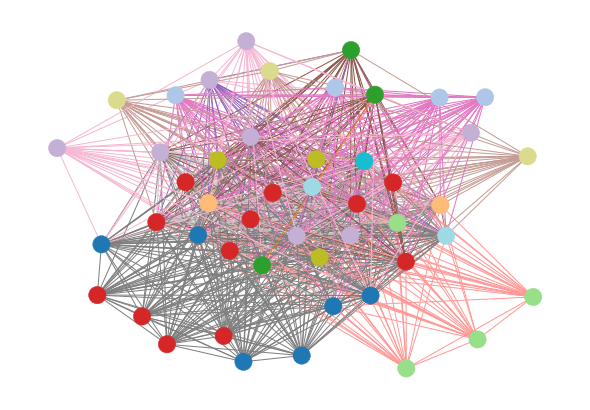}}}
\caption{Homogeneous Graph Examples using SCG dataset.} \label{fig:Graph12}
\end{figure}

\section{SCG Dataset}
Here, we discuss the development of the SCG dataset, detailing the processes of dataset collection, curation, feature selection, and exploratory statistics.

\subsection{Data Collection}
Data was collected from the central database system of one of the largest and most prominent FMCG (Fast Moving Consumer Goods) companies in Bangladesh. The dataset was reorganized for temporal graph utilization, extracting relevant nodes and features. The actual product names, product codes, and the name of the company are not disclosed to maintain the competitive standing of the concerned company. Table \ref{tab:datasets}  summarizes the dataset information.

Here, we provide information on the quality control of the dataset, its coverage, and availability.

\textbf{Quality Control.} Each individual node, edge, and node feature undergoes a thorough manual examination and validation process. This involves scrutinizing the data for any irregularities, such as anomalies or missing information. Additionally, even zero values are meticulously reviewed to ensure their accuracy and legitimacy. 

\textbf{Temporal \& Location Coverage.} The occurrences in this database cover the period from January 1, 2023, to August 9, 2023, and include four features: production, sales orders, delivery to distributors, and factory issues in two metrics: unit and weight produced in metric tons. The company operates across the entirety of Bangladesh.

\textbf{Dataset Availability.} . The dataset is publicly avail-
able on GitHub at \textit{\href{https://github.com/CIOL-SUST/SCG}{https://github.com/CIOL-SUST/SCG}} under the LGPL-2.1 Licence. 

\textbf{Graphs in Dataset.} The dataset contains graphs in both homogeneous (see examples in Figure \ref{fig:Graph12}) and heterogeneous (see example in Figure \ref{fig:heterogeneous-graph}) formats, with four temporal data points as node features. The construction, implementation, feature selection, and use of different graph properties depend on the specific application and requirements. In this paper, we provide a brief description of these aspects.



\begin{figure}
\centering
\subfigure[Number of products in each product group.]{%
\resizebox*{7cm}{!}{\includegraphics[scale=0.35]{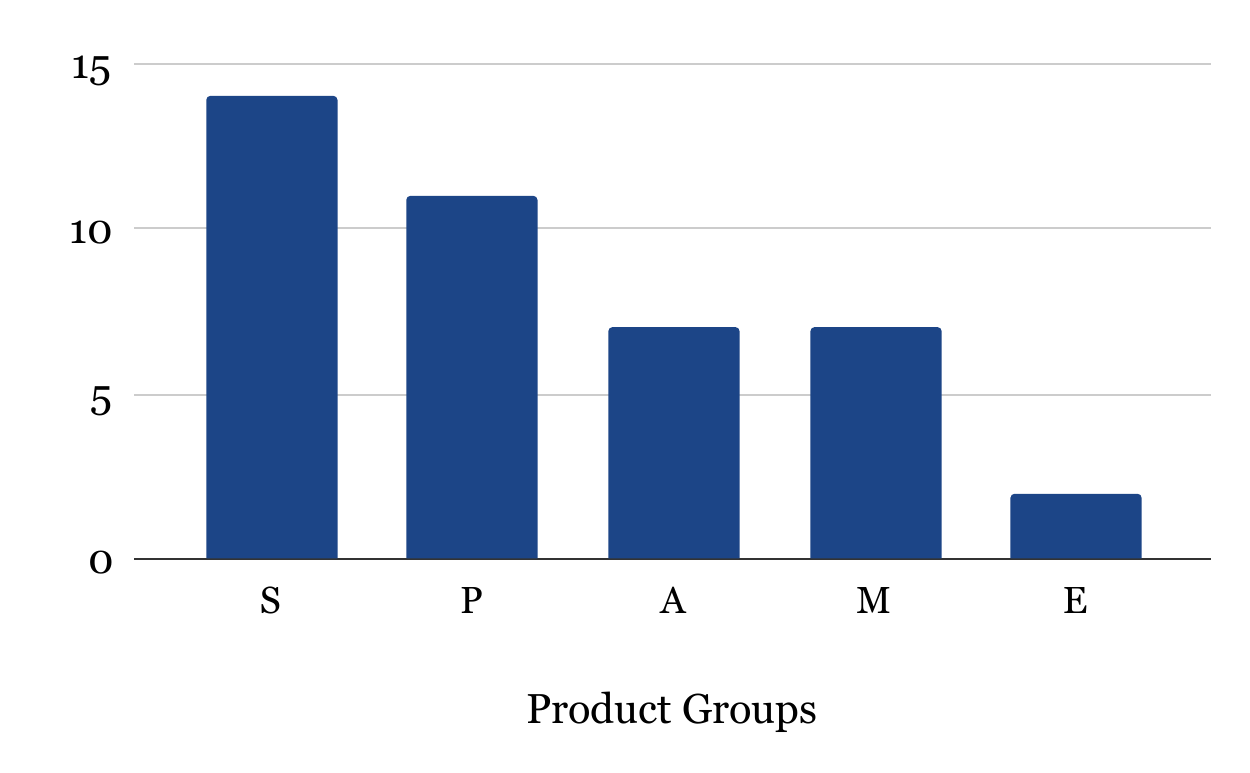}}}\hspace{5pt}
\subfigure[Number of products in each product sub-group.]{%
\resizebox*{7cm}{!}{\includegraphics[scale=0.35]{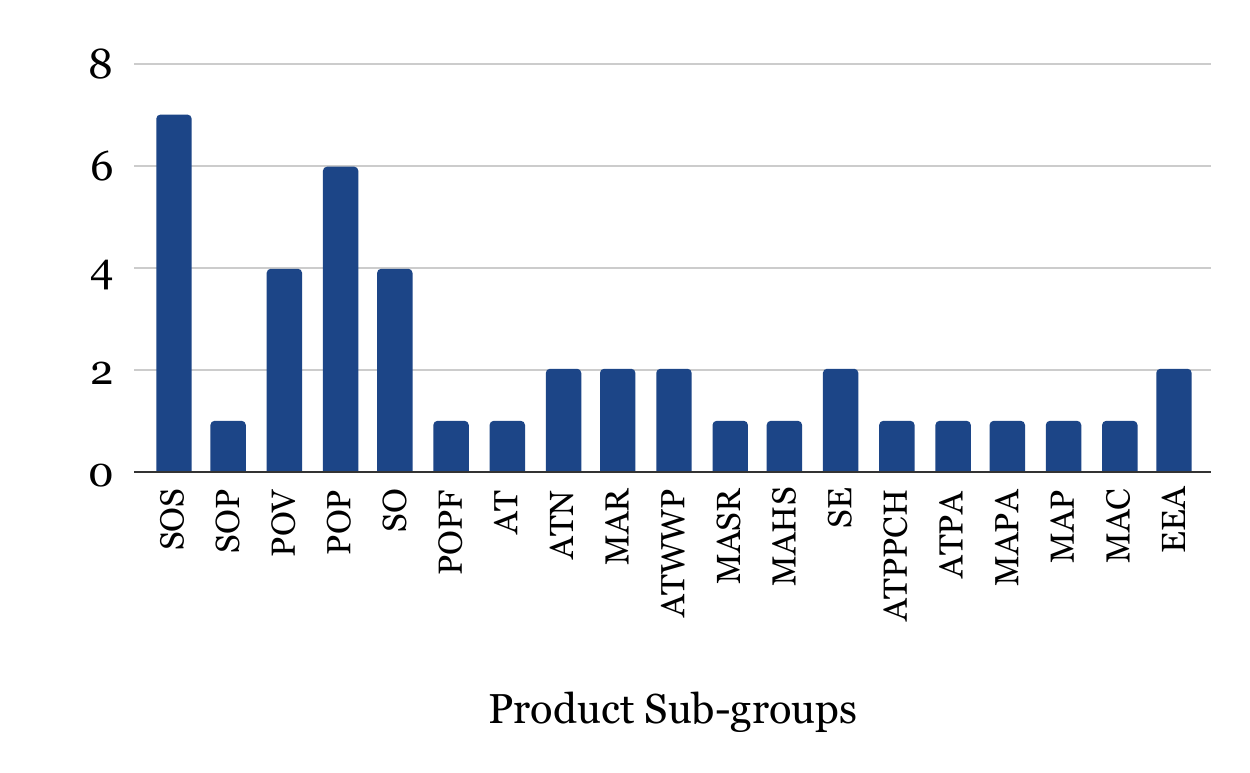}}}
\caption{Dataset Statistics.} \label{fig:PG/SPG-Count}
\end{figure}

\begin{table}
\tbl{Dataset Edges Information.}
{\begin{tabular}{lrlr} 
\toprule
 \multicolumn{2}{c}{ Edge Classes }  & \multicolumn{2}{c}{ Edge Count } \\
 \midrule
Total Edge Types   & 62   & Total Unique Edges  & 684 \\
Class (Group)      & 5    & Count (Group)      & 188\\
Class (Sub-group)  & 19   & Count (Sub-group)  & 52\\
Class (Plant)      & 25   & Count (Plant)      & 1647\\
Class (Storage)    & 13   & Count (Storage)    & 3046\\
 \bottomrule
\end{tabular}}
\label{tab:datasets}
\end{table}

\subsection{Data Feature Description}
Within SCG, \textbf{nodes} pertains to distinct products, while \textbf{edges} represent various connections linking these products: the same product group or subgroup, the same plant or storage location. Figure \ref{fig:Graph12} presents some examples of homogeneous graphs.

In the temporal data, node features include production, sales orders, delivery to distributors, and factory issues. We have all the temporal data in two modes: number of units produced (example: 500 units) and total weight (example: 10 metric tons) produced.

Definition of temporal features:
\begin{itemize}
\item \textit{Production}, which quantifies product output considering sales orders, customer demand, vehicle fill rate, and delivery urgency. This quantity is typically measured in units or Metric Tons. For example, a production value might be 500 units or 10 metric tons produced in a given timeframe.
\item \textit{Sales Order} signifies distributor-requested quantities, pending approval from the accounts department. It reflects overall product demand. For example, a sales order might indicate a request for 300 units or 6 metric tons.
\item \textit{Delivery to Distributors} denotes dispatched products aligning with orders, impacting company revenue significantly. For example, delivery data might show that 450 units or 9 metric tons have been sent to distributors.
\item \textit{Factory Issue} covers total products shipped from manufacturing facilities, with some going to distributors and the rest to storage warehouses. For example, factory issues might report that 700 units or 14 metric tons have been dispatched, with 500 units going to distributors and the remaining 200 units going to warehouses.
\end{itemize}

\subsection{Data Statistics}
\subsubsection{Product Group and Sub-groups}

Figure \ref{fig:PG/SPG-Count}(a) shows that product group “S” has the highest number of products, reflecting its highest variety, and product group “E” has the lowest number of products. Notably, a majority of the product groups exhibit a range of around 8 to 10 products each.
Figure \ref{fig:PG/SPG-Count}(b) reveals the diversity within product subgroups. As previously noted, the "s" product group stands out for its remarkable variety, and this characteristic is consistently maintained within its corresponding subgroup. Among our product subgroups, "SOS" emerges as the one encompassing the largest array of product categories. Interestingly, despite the elevated diversity exhibited by the product group, the corresponding subgroup maintains a relatively stable count.


\begin{figure}
\centering
\subfigure[All product's production (SKU) of 'POV' sub-group.]{%
\resizebox*{7cm}{!}{\includegraphics[scale=0.5]{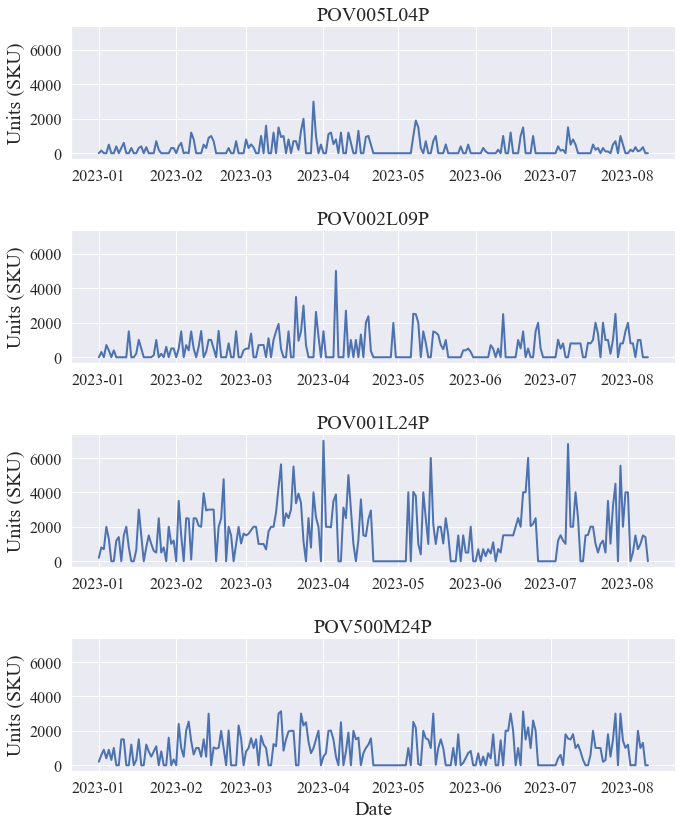}}}\hspace{5pt}
\subfigure[All 04 temporal values (SKU) of product 'SOS008L02P'.]{%
\resizebox*{7cm}{!}{\includegraphics[scale=0.5]{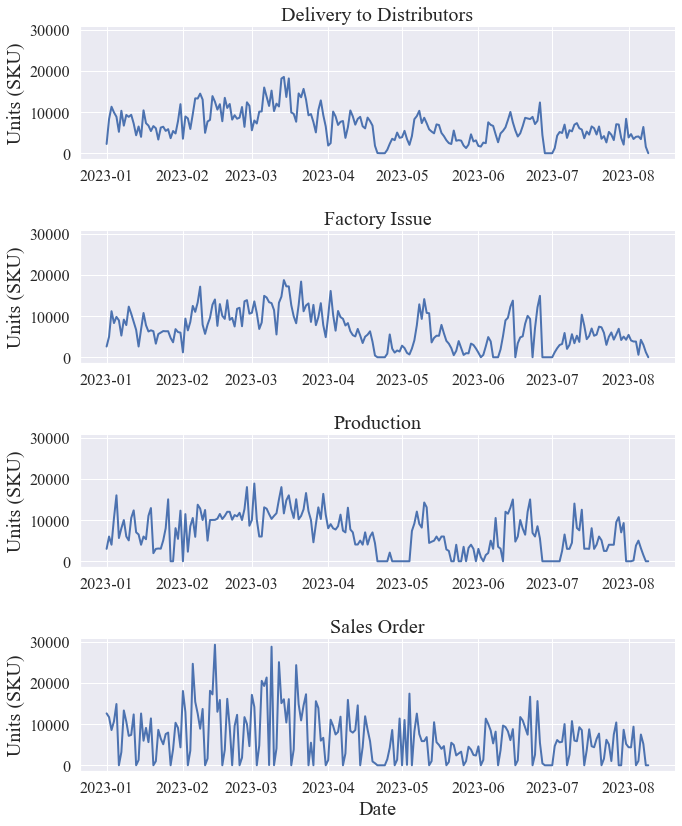}}}
\caption{Data Samples.} \label{fig:DataSamples}
\end{figure}

\subsubsection{Analyzing Temporal Trends.}
Figure \ref{fig:DataSamples} (a), A particularly attractive pattern emerges: the production curve exhibits distinct and sharp fluctuations. Notably, the production unit of the company occasionally closes, which can be attributed to factors such as scheduled holidays, adjustments in the product's MRP rate, demand variations, and optimizations in transportation policies. In contrast, the other aspects of the graph, encompassing sales orders, delivery to distributors, and factory issuance, remain notably stable and consistent. Their minimal deviations stand in contrast to the variabilities characterizing the production curve, providing an insight into the dynamics of the supply chain activities.

In, Figure \ref{fig:DataSamples} (b), we can notice a very interesting pattern. The product is manufactured in large batches, with orders being accumulated over several days before production begins. The processes of delivering to distributors, handling factory issues, and fulfilling sales orders appear to be running smoothly and consistently. The company has opted to operate the production plant intermittently and maintain product inventory. When you observe straight lines in the production graph, it indicates that the plant is not actively engaged in production during those periods.

\subsubsection{Analysing Temporal Correlations.}
\begin{figure}
\centering
\subfigure[Temporal correlation of some products from Product Group A.]{%
\resizebox*{7cm}{!}{\includegraphics[scale=0.6]{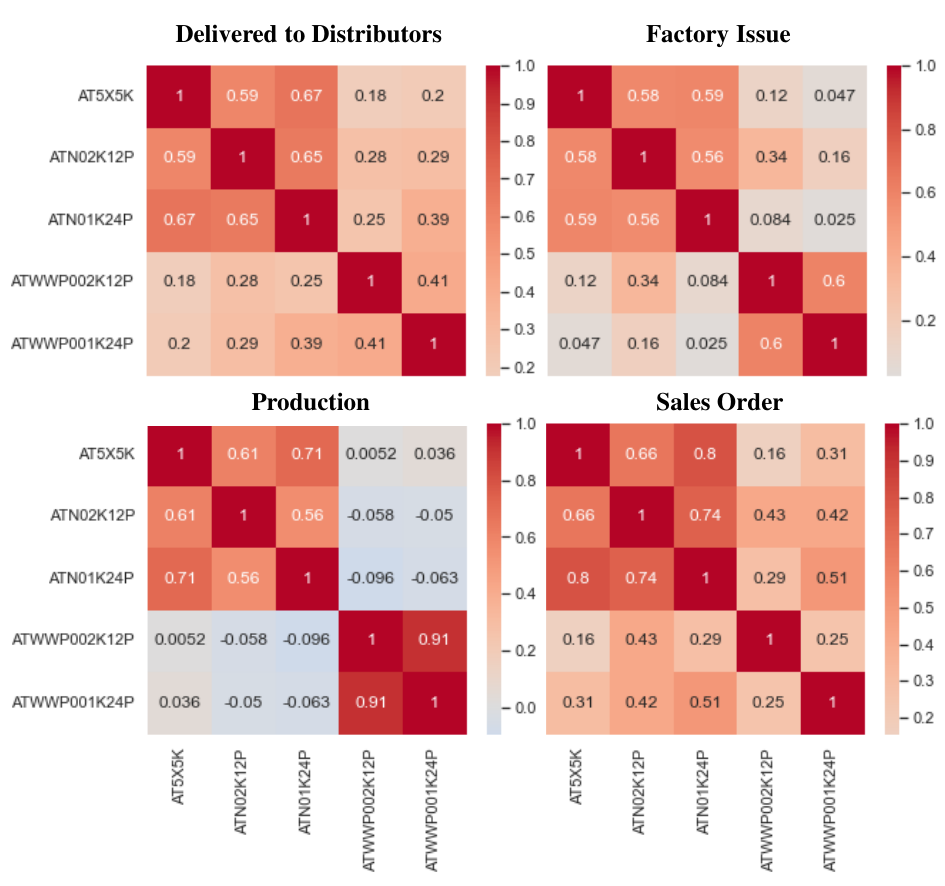}}}\hspace{5pt}
\subfigure[Temporal correlation of some products from Product Group A.]{%
\resizebox*{7cm}{!}{\includegraphics[scale=0.6]{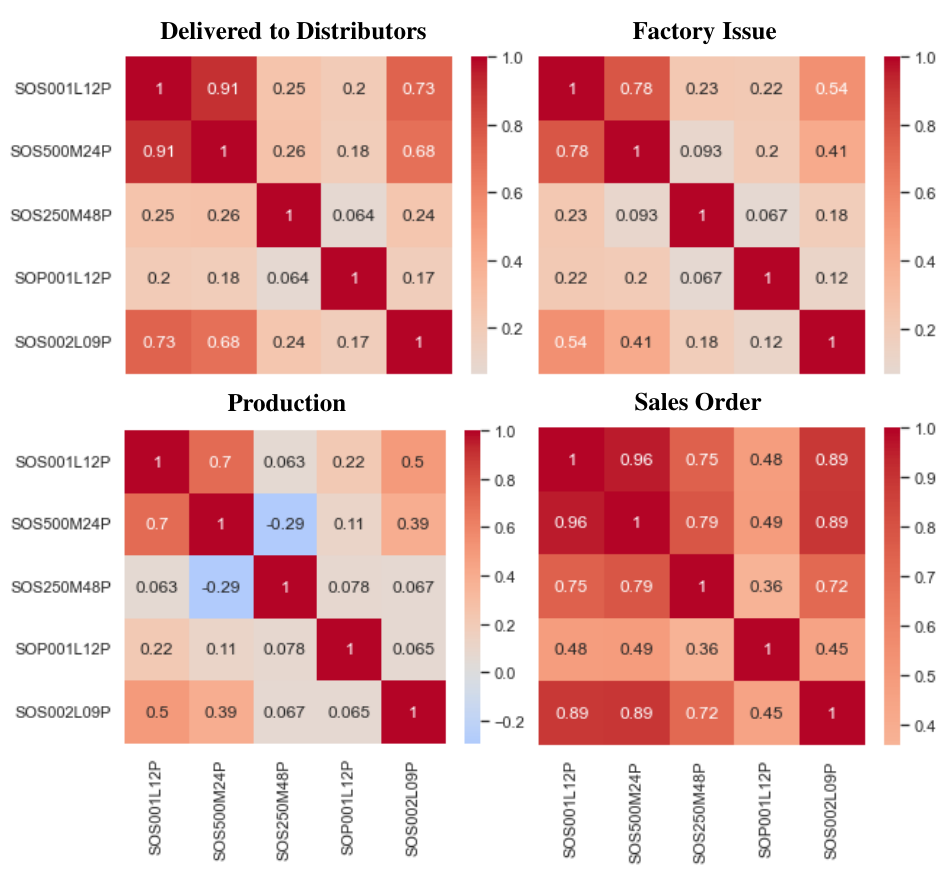}}}
\caption{Temporal correlations.} \label{fig:G-A-Corrs}
\end{figure}

The correlation plot portrayed in Figure \ref{fig:G-A-Corrs}(a) serves to demonstrate the ways in which diverse subgroups of product "A" are interconnected in a multitude of facets. It is worth noting that the relationship between delivery to distributors and sales orders is particularly robust, whereas the production unit exhibits a distinct and prominent correlation pattern. Furthermore, a noteworthy observation pertains to the correlation between the production quantities of two subgroups, specifically "ATWWPOO2K12P" and "ATWWPOO1K24P". Additionally, it is worth highlighting that factory issues and sales orders similarly display correlation trends, albeit with some discernible variations.


Upon examining the correlation plot of the "S" subgroup in Figure \ref{fig:G-A-Corrs}(b), a distinct pattern becomes apparent: there is a significant and consistent correlation between the "SOSOO1L12P" and "SOS500M24P" subgroups across all variables. Notably, an intriguing observation arises in the context of sales orders, where a majority of the products within the "S" subgroup demonstrate a relatively higher correlation compared to delivery to distributors, factory issuances, and production.

The comprehensive analysis of correlations and temporal activities strongly indicates a substantial opportunity for the application of graph-based machine learning methodologies in the realm of supply chains. Leveraging the inherent structures and relationships within supply chain networks, graph-based machine learning holds great promise in enhancing decision-making processes, optimizing resource allocation, and fostering adaptability in the dynamic landscape of supply chain operations. 

\section{Practical Applications of Supply Chain as Graphs}
In this section, we explore the practical applications of representing supply chains as graphs. We illustrate how graph-based approaches can optimize various supply chain functions and improve overall efficiency and decision-making. 

\subsection{Homogeneous Graphs} 
In homogeneous form, our SCG dataset's structural design models products as nodes, while the interrelationships—such as shared product groups, subgroups, production facilities, and storage locations—are represented as edge connections. This setup enables the application of GNNs to tackle complex supply chain issues, such as: 

\begin{itemize}
    \item  \textbf{Supply Chain Planning} involves the utilization of historical data pertaining to production and demand as node attributes, with the integration of time-related features to capture seasonality and trends. To enhance demand forecasting accuracy, related entities' influence can be harnessed through neighbourhood nodes. The hierarchical structure of nodes, including product groups, subgroups, plants, and storage locations, can be used for hierarchy-aware forecasting. In data-scarce scenarios, improved forecasting can be achieved by applying transfer learning across nodes of the same classification.
    
    \item  \textbf{Product Classification} involves classifying products by product groups, subgroups, facilities; and also by economic profitability and production similarities for economic decision-making.
    \item  \textbf{Product Relation Classification} involves classifying or predicting product relations in the supply chain graph. Product Relation detection involves detecting a missing edge or relation in the supply chain graph by a binary prediction objective.
    \item  \textbf{Supply Chain Optimization} involves modelling goods flow between nodes, considering demand, lead times, and reorder points. Optimal routes and quantities can be recommended using edges as transportation links. Nodes representing plants can also aid in suggesting production adjustments based on demand projections, constraints, and capacity.
    \item  \textbf{Anomaly Detection} is achieved by contrasting predicted demand with actual data to identify disruptions, stock outs, or unusual demand spikes. Deviations in demand from aggregated neighbourhood demand can signal inconsistencies within neighbourhoods, which can be detected as anomalies.
    \item  \textbf{Event Classification} involves the training of GNNs to classify events based on edges and nodes, such as production capacity changes, new product launches, and disruptions. 
    \item  \textbf{Fluctuation Detection} entails training GNNs to identify and classify fluctuations in the network based on impacts of global price hikes, supply chain issues, and disruptions. Fluctuations in supply chains refer to disruptions that can occur due to various factors and can significantly impact the operations of businesses. These disruptions can amplify negative shocks, affecting not just the firm experiencing the failure, but also its suppliers and customers, and even firms in other parts of the production network \citep{NBERw27565}. Using temporal data, the model can recognize patterns in these fluctuations over time and determine if any demand and supply fluctuation can occur in the near future for better planning. 
    \item  \textbf{Combinatorial} \citep{cappart2022combinatorial} and \textbf{Constrained} \citep{kotary2021endtoend} \textbf{Optimization} involves utilizing GNNs to optimize complex decision-making within the supply chain, considering various factors and constraints. GNNs can effectively model the intricate relationships between nodes and edges, enabling efficient allocation of resources, route planning, and inventory management, ultimately leading to improved  performance.
\end{itemize}

\subsection{Heterogeneous Graphs} 
All these tasks discussed can be executed using both heterogeneous and homogeneous graphs, but heterogeneous graphs excel in modelling complex systems. By incorporating diverse nodes (e.g., products, storage locations, production facilities) and edges representing their intricate relationships, heterogeneous graphs capture the complex interactions within supply chain networks. This approach enables a comprehensive representation of varying supply chain components and their interdependencies, addressing challenges and enhancing decision-making. Ultimately, it provides deeper insights into supply chain dynamics, allowing for optimized strategies and improved performance and resilience.

\subsection{Hypergraphs} 
A hypergraph generalizes a graph by allowing edges, called hyperedges, to connect multiple vertices, rather than just two. This capability is useful for modelling complex relationships where interactions involve several entities. In supply chain management, hypergraphs can represent intricate relationships among products, storage locations, and production facilities through hyperedges. This approach captures the complexity of supply chain networks more effectively, offering a dynamic framework to manage the diverse characteristics and interdependencies of supply chain components. Utilizing hypergraphs can enhance understanding, performance, and resilience within supply chain management.

\section{Experiments and Benchmarking}
Here, we model our data with several benchmark graph-based models using PyTorch Geometric and PyTorch Geometric Temporal libraries \citep{rozemberczki2021pytorch}.  We use plant connections as edges in the graph to get predictions for all products, together with 8:2 train-test split of the temporal dataset for forecasting, planning, and anomaly detection. In product classification, product relation detection, and classification, demands are used as node features. Product classification encompasses the significant task of categorizing products, while edge classification, or detection, involves discerning storage relationships. Our models underwent training for 200 epochs, with the possibility of early stopping if they converged ahead of time. Different metrics were employed depending on the specific task. It's important to note that we included non-neural network-based methods in our comparative analysis.

\subsection{Experimental Details}  \label{app-exp-mp}
Here, we discuss all the experimental details of our bench-marking process for each task on both homogeneous and heterogeneous graphs.

\subsubsection{Supply Chain Planning (Demand) and Forecasting}
In these tasks, we use a temporal approach where the node features display temporal and dynamic characteristics, while nodes and edges maintain static attributes to predict demand (Sales Order). The edges represent the similarity between different plants or production facilities, and the products are represented as nodes. In the case of traditional models, we independently model each product and subsequently aggregate the results. Conversely, with temporal GNNs, all predictions are generated simultaneously. It's worth noting that we have maintained the use of default hyperparameters, as mentioned previously. MSE loss is used.
Among the models, ARMA \citep{Reinsel1993523453453} is a traditional statistical model, 
GRU \citep{chung2014empirical} and  LSTM \citep{hochreiter1997long} are deep learning based models,
DCRNN \citep{li2018diffusion}, TGCN \citep{Zhao2018TemporalGC}, GConvGRU \citep{seo2016structured}  are GNN-based models. In heterogeneous GNNs, we use THGNN \citep{Xiafafang_2022}, HTGNN \citep{fan2021heterogeneoustemporalgraphneural} and HSTGCN \citep{9546afaefaf488} with default settings.
Our approach for production forecasting model is the same to demand prediction model, with the distinction being that the primary output objective was production rather than demand.

\subsubsection{Product Classification}
In the context of product classification, we harnessed the temporal features as static node attributes to forecast the product group, with the edges denoting the similarities between production facilities. The loss function was cross-entropy loss. Our Artificial Neural Network (ANN) model featured two linear layers, the K-Nearest Neighbors (KNN) algorithm incorporated a setting of 5 closest neighbours, and the Graph Attention Network (GAT) employed 4 attention heads with 3 layers in each GNN.
Among the models, Logistic Regression is a statistical model, XGB Classifier \citep{Chen:2016:XST:2939672.2939785} is also a traditional boosting based model, KNN \citep{Mucherino2009knn} is a statistical clustering based model, ANN is a deep learning based model, GCN \citep{kipf2017semisupervised} and GAT \citep{Velickovic:2018we}  are GNN-based models. In the heterogeneous task, we used temporal features as static node attributes to predict product groups, where edges represented similarities between production facilities. 

\subsubsection{Anomaly Detection}
For anomaly detection, we employed sales orders as dynamic temporal node characteristics, applying a model with MSE loss to identify anomalies within time series data. Increased data fluctuations across the graph during a specific time frame are anomaly as an anomaly and warrant communication with the company. It is a supervised approach; following GDN \citep{GDN-xxx}. Among the models,  PCA is a statistical model;
ANN, AE \citep{Aggarwal2015w4525} and LSTM-VAE \citep{park2017multimodal} are deep learning based models;
GANF \citep{GANF} and GDN \citep{GDN-xxx} are GNN-based temporal anomaly detection models.
In the heterogeneous task, we utilized sales order data as dynamic temporal node characteristics. Significant data fluctuations across the graph within a specific time frame are flagged as anomalies, requiring communication with the company. Ad-EHG \citep{8276segewagt899}  and HRGCN \citep{li2023hrgcn} are the heterogeneous GNN-based models here.

\subsubsection{Product Relationship Detection and Classification}
Product relation detection entailed a binary classification task focused on discerning the presence or absence of edges, while product relation classification involved a multi-class classification task. The loss function utilized in this context was the cross-entropy loss. As before, the ANN model featured two linear layers, the KNN algorithm incorporated a setting of 5 closest neighbours, and the GAT employed 4 attention heads with 3 layers in each GNN. In the heterogeneous task, it is also designed as a binary classification objective aimed at distinguishing the presence or absence of edges. In contrast, the product relation classification involved a multi-class classification goal. The loss function employed here was cross-entropy loss.

\begin{table}
\tbl{Benchmark scores and Comparison of Various Models on Different Tasks.}{
\begin{tabular}{cccccc}
\toprule
\multicolumn{2}{c}{\textbf{Models and Type}} & \multicolumn{2}{c}{\textbf{(a) SC Planning (Demand) \textsuperscript{Hm}}} & \multicolumn{2}{c}{\textbf{(b) SC Planning (Demand) \textsuperscript{Ht}}} \\
\cmidrule(r){1-2} \cmidrule(r){3-4} \cmidrule(r){5-6}
Type & Model & RMSE ($\downarrow$) & Mean Diff. ($\downarrow$) & RMSE ($\downarrow$) & Mean Diff. ($\downarrow$) \\
\midrule
Statistical & ARMA & 37.9875 & 486.8858 &  37.7776 & 486.6877\\
Deep Learning & GRU & 33.2277 & 359.8285 &  32.3457 & 358.7857\\
Deep Learning & LSTM & 30.7782 & 304.6777 &  30.2656 & 304.1456\\
GNN-based & DCRNN\textsuperscript{Hm}/THGNN\textsuperscript{Ht} & \textbf{27.9811} & \textbf{253.1135} & 28.1633 & 253.3335 \\
GNN-based & TGCN\textsuperscript{Hm}/HTGNN\textsuperscript{Ht} & 28.0145 & 253.9384 & \textbf{26.2424} & \textbf{251.2342} \\
GNN-based & GConvGRU\textsuperscript{Hm}/HSTGCN\textsuperscript{Ht} & 28.1814 & 272.6487 & 29.2522 & 281.3533\\
\midrule

\multicolumn{2}{c}{\textbf{Models and Type}} & \multicolumn{2}{c}{\textbf{(c) Production Forecasting \textsuperscript{Hm}}} & \multicolumn{2}{c}{\textbf{(d) Production Forecasting \textsuperscript{Ht}}} \\
\cmidrule(r){1-2} \cmidrule(r){3-4} \cmidrule(r){5-6}
Type & Model & RMSE ($\downarrow$) & Mean Diff. ($\downarrow$) & RMSE ($\downarrow$) & Mean Diff. ($\downarrow$) \\
\midrule
Statistical & ARMA  & 37.4542 & 483.5387  & 37.1332 & 480.2552\\
Deep Learning & GRU  & 32.3532 & 354.2312  & 33.5222 & 356.2322\\
Deep Learning & LSTM  & 30.2142 & 300.2942 & 30.1322 & 297.3422\\
GNN-based & DCRNN\textsuperscript{Hm}/THGNN\textsuperscript{Ht} & 27.9434 & \textbf{252.6459} & 28.4745 & 256.6222 \\
GNN-based & TGCN\textsuperscript{Hm}/HTGNN\textsuperscript{Ht}  & \textbf{27.9244} & 253.2390 & \textbf{27.1232} & \textbf{249.3252} \\
GNN-based & GConvGRU\textsuperscript{Hm}/HSTGCN\textsuperscript{Ht} & 28.0414 & 270.5629  & 29.9235 & 276.3662 \\
\midrule

\multicolumn{2}{c}{\textbf{Models and Type}}  & \multicolumn{2}{c}{\textbf{(e) Product Cls. \textsuperscript{Hm}}} & \multicolumn{2}{c}{\textbf{(f) Product Cls. \textsuperscript{Ht}}} \\
\cmidrule(r){1-2} \cmidrule(r){3-4} \cmidrule(r){5-6}
Type & Model & Accuracy & Precision & Accuracy & Precision \\
\midrule
Statistical & Logistic Reg. & 66.67\% & 76.67\% & 69.32\% & 73.24\% \\
Boosting-based & XGB & 65.56\% & 61.48\% & 71.45\% & 61.42\% \\
Statistical & KNN & 64.44\% & 64.07\% & 72.23\% & 74.07\% \\
Deep Learning & ANN & 66.37\% & 67.78\% & 74.67\% & 75.18\% \\
GNN-based & GCN & 71.23\% & 72.18\% & 81.35\% & 78.34\% \\
GNN-based & GAT & \textbf{75.68\%} & \textbf{75.32\%} & \textbf{82.24\%} & \textbf{79.56\%} \\
\midrule

\multicolumn{2}{c}{\textbf{Models and Type}}  & \multicolumn{2}{c}{\textbf{(g) Product Relation Cls. \textsuperscript{Hm}}} & \multicolumn{2}{c}{\textbf{(h) Product Relation Cls. \textsuperscript{Ht}}} \\
\cmidrule(r){1-2} \cmidrule(r){3-4} \cmidrule(r){5-6}
Type & Model & Precision & Recall & Accuracy & Precision \\
\midrule
Statistical & Logistic Reg. & 62.73\% & 66.78\% & 66.12\% & 67.18\% \\
Boosting-based & XGB & 71.78\% & 71.86\% & 71.22\% & 71.13\% \\
Statistical & KNN & 74.72\% & 74.75\% & 74.23\% & 73.73\% \\
Deep Learning & ANN & 76.58\% & 77.72\% & 77.23\% & 77.79\% \\
GNN-based & GCN & \textbf{91.36\%} & \textbf{82.27\%} & 88.23\% & 87.14\% \\
GNN-based & GAT & 89.43\% & 81.22\% & \textbf{90.18\%} & \textbf{91.02\%} \\
\midrule

\multicolumn{2}{c}{\textbf{Models and Type}}  & \multicolumn{2}{c}{\textbf{(i) Product Relation Det. \textsuperscript{Hm}}} & \multicolumn{2}{c}{\textbf{(j) Product Relation Det. \textsuperscript{Ht}}} \\
\cmidrule(r){1-2} \cmidrule(r){3-4} \cmidrule(r){5-6}
Type & Model & Accuracy & Precision & Accuracy & Precision \\
\midrule
Statistical & Logistic Reg. & 68.63\% & 75.15\% & 67.24\% & 68.37\% \\
Boosting-based & XGB & 71.23\% & 81.12\% & 71.25\% & 75.32\% \\
Statistical & KNN & 74.63\% & 78.26\% & 75.38\% & 75.73\% \\
Deep Learning & ANN & 78.74\% & 81.62\% & 78.26\% & 80.15\% \\
GNN-based & GCN & 89.32\% & 88.28\% & \textbf{92.12\%} & 89.51\% \\
GNN-based & GAT & \textbf{91.45\%} & \textbf{91.64\%} & 91.27\% & \textbf{90.15\%} \\
\midrule

\multicolumn{2}{c}{\textbf{Models and Type}} & \multicolumn{2}{c}{\textbf{(k) Anomaly Det.\textsuperscript{Hm}}} & \multicolumn{2}{c}{\textbf{(l) Anomaly Det.\textsuperscript{Ht}}} \\
\cmidrule(r){1-2} \cmidrule(r){3-4} \cmidrule(r){5-6}
Type & Model & Precision &   Recall  & Precision &   Recall \\
\midrule
Statistical & PCA &  31.33\%  & 27.34\%  &  27.24\%  & 28.43\% \\
Deep Learning & ANN  & 55.24\%  & 47.75\% & 53.43\%  & 49.24\% \\
Deep Learning  & AE   & 75.78\%  &  71.78\% & 71.22\%  &  69.32\%\\
Deep Learning & LSTM-VAE   & 91.87\% & 86.58\%   & 89.23\% & 87.24\%  \\
GNN-based & GANF\textsuperscript{Hm}/Ad-EHG\textsuperscript{Ht}   & 85.45\% & 76.33\%   & 90.34\% & 89.13\% \\
GNN-based & GDN\textsuperscript{Hm}/HRGCN\textsuperscript{Ht} & \textbf{94.15\%} & \textbf{87.76\%} &   \textbf{92.43\%} & \textbf{89.34\%} \\
\bottomrule
\end{tabular}}
\tabnote{\textsuperscript{Hm}Homogeneous Graph modelling, \textsuperscript{Ht}Heterogeneous Graph modelling, Cls. = Classification, Det. = Detection}
\label{tab:combined-modelling}
\end{table}

\subsection{Results} 
Table \ref{tab:combined-modelling} reveals that graph-based models consistently achieve higher scores compared to other models in the experiments. The experiments were performed multiple times and an average is taken.

\subsubsection{Supply Chain Planning (Demand)} 
In demand prediction (see Table \ref{tab:combined-modelling}(a)), we observe a range of models, including traditional time series methods like ARMA and more advanced recurrent neural networks such as GRU and LSTM. Notably, the newer graph-based Graph Convolutional GRU (GConvGRU) achieves competitive performance. It's evident that the recurrent models, particularly DCRNN and TGCN, outperform ARMA, demonstrating the benefits of leveraging sequential information with GNNs. GNN-based models like GConvGRU also showcase their potential in modelling supply chain demand patterns. Heterogeneous graph segment also follows the same pattern.

\subsubsection{Production Forecasting} 
Similar to demand prediction (see Table \ref{tab:combined-modelling}(c)), production forecasting shows a similar trend with respect to model performance. The recurrent models, DCRNN and TGCN, continue to outperform ARMA, emphasizing the superiority of GNN-based methods for this task. LSTM stands out as the best-performing individual model, closely followed by DCRNN and TGCN, signifying their utility in forecasting production levels using GNNs. GNNs also outperforms other models in heterogeneous setup, too.

\subsubsection{Product Classification} 
In product classification demonstrated in Table \ref{tab:combined-modelling}(e), traditional logistic regression achieves a respectable accuracy of 66.67\%, but GNN-based models like GCN and GAT exhibit competitive performance with an accuracy of 71.23\% and 75.68\%, respectively. The performance of these models showcases their adaptability for supply chain product classification tasks.

In heterogeneous product classification (Table \ref{tab:combined-modelling}(f)), we observe improved performance compared to the homogeneous version. Traditional logistic regression achieves 69.\% accuracy, while GNN-based models like heterogeneous GCN and GAT demonstrate competitive performance, achieving accuracies of 81.35\% and 82.24\%, respectively; showing effectiveness of GNN models.

\subsubsection{Product Relation Classification} In product relation classification demonstrated in Table \ref{tab:combined-modelling}(g), showcases the efficiency of GNNs, with GCN and GAT achieving remarkable accuracy levels of 92.12\% and 91.27\%, respectively; better than homogeneous counterparts. They significantly outperform traditional methods and even other machine learning algorithms, demonstrating their prowess in classifying complex product relationships.

In product relation classification for heterogeneous graphs (Table \ref{tab:combined-modelling}(h)), GNNs like GCN and GAT achieve notable accuracy levels of 88.23\% and 90.18\%, respectively, though slightly lower than their homogeneous counterparts. But, these models excel over traditional methods and other machine learning algorithms, demonstrating their efficacy in classifying intricate product relationships.

\subsubsection{Product Relation Detection} 
The task of detecting relationships between products is approached effectively by various models. Table \ref{tab:combined-modelling}(i) shows that, GCN and GAT outperform the traditional approaches, achieving an accuracy of 89.32\% and 91.45\%, respectively. These results underscore the capacity of GNNs to model intricate product relationships in the supply chain.

In product relation detection for heterogeneous graphs (Table \ref{tab:combined-modelling}(j)), GNNs such as GCN and GAT achieve impressive accuracy levels of 92.12\% and 91.27\%, respectively, surpassing their homogeneous counterparts. They outperform traditional methods and other machine learning algorithms, showcasing their effectiveness in classifying complex product relationships.
    
\subsubsection{Anomaly Detection} 
Anomaly detection involves identifying deviations from expected patterns in time series. In this task, Table \ref{tab:combined-modelling}(k) shows that LSTM-VAE and GDN lead the way in terms of precision and recall, underlining their effectiveness in pinpointing anomalies. The GANF model also demonstrates strong performance, emphasizing the potential of GNN-based approaches for anomaly detection.

In heterogeneous anomaly detection, the goal is to detect deviations from expected patterns in dynamic heterogeneous graphs. Table \ref{tab:combined-modelling}(l) highlights that GNN-based HRGCN \citep{li2023hrgcn} and Ad-EHG \citep{8276segewagt899} achieve superior precision and recall, showcasing their effectiveness in identifying anomalies and highlighting the potential of GNN-based methods for anomaly detection.

\subsubsection{Comparison between Homogeneous and Heterogeneous Tasks} 
The benchmark scores presented in Table \ref{tab:combined-modelling} reveal that GNN-based models consistently outperform statistical and deep learning models across all tasks. Comparing heterogeneous and homogeneous graph modelling, heterogeneous models generally achieve better performance on average. For Supply Chain planning and production forecasting, GNN-based models like TGCN and HTGNN show lower RMSE and Mean Diff. in the heterogeneous versions. In product classification and relation classification, GNN-based models (GAT) achieve higher accuracy and precision in the heterogeneous setting. Similarly, for product relation detection and anomaly detection, heterogeneous models (GAT and HRGCN) outperform their homogeneous counterparts. This suggests that incorporating heterogeneous graph structures can capture more complex relationships, leading to improved predictive accuracy and anomaly detection capabilities.

\section{Discussion} 
We believe this dataset can be a foundational cornerstone for using GNNs in supply chain analysis and problem-solving. The current landscape suffers from a lack of datasets tailored to the complexities of supply chain management. Our contribution addresses this gap by providing a comprehensive benchmark graph dataset specifically designed for supply chain applications, covering various tasks and challenges. This dataset enables the modelling of intricate network relationships, leading to more robust decision-making and optimization. By establishing standardized evaluation metrics and methodologies, we lay the groundwork for solving supply chain problems using GNNs, promoting advancements and innovation in the field.

\vspace{2mm}

\textbf{Societal Impact.} \quad
The societal impact of our work is significant. By improving supply chain analysis and optimization through GNNs, we can enhance efficiency in critical areas such as global trade, manufacturing, and logistics. This can lead to more reliable and resilient supply chains, reducing disruptions and ensuring timely delivery of goods. Enhanced supply chain management also has implications for environmental sustainability, as optimized logistics can reduce waste and lower carbon footprints. Furthermore, our dataset and methodologies can spur further research and development in both academia and industry, fostering innovation and creating new opportunities for economic growth. By addressing the current gaps in data and methodology, our work supports more informed decision-making and drives progress towards more sustainable and efficient supply chain practices.

\vspace{2mm}
\textbf{Managerial Insights for Decision Makers in Industry.} \quad
Our research provides valuable insights for decision-makers in the industry. By leveraging GNNs, managers can gain a deeper understanding of complex supply chain dynamics and make data-driven decisions that enhance efficiency and resilience. The benchmark dataset enables precise modelling of supply chain networks, facilitating better risk management and identification of hidden dependencies. Additionally, the improved accuracy in forecasting and anomaly detection aids in proactive problem-solving and resource optimization. Ultimately, adopting GNN-based solutions can lead to significant cost savings and improved operational performance.

\vspace{2mm}

\textbf{Limitations.} \quad
We acknowledge that the dataset's temporal scope is limited, which constrains its ability to capture long-term supply chain dynamics. However, its concise timeframe is well-suited for analysing immediate responses and short-term intricacies in supply chain management. Despite this limitation, the dataset remains a valuable tool for addressing diverse supply chain challenges. The utility of the dataset depends on the chosen model or problem and varies based on how different information is represented as nodes or edges. It is up to researchers to determine the most effective way to leverage its potential. 

\vspace{2mm}
\textbf{Future Research Directions.} \quad
Future research directions in this field could explore the integration of more diverse and granular data sources to enhance the robustness and applicability of GNN models in supply chain management. Expanding the temporal scope and incorporating real-time data can provide deeper insights into long-term trends and immediate responses. Additionally, investigating the potential of heterogeneous modelling by including varied types of nodes, such as storage locations and production facilities, could lead to more comprehensive and impactful analyses. Exploring the intersection of GNNs with other advanced technologies, like reinforcement learning and AI-driven optimization, could further elevate decision-making processes. Finally, developing standardized benchmarks and evaluation metrics will be crucial for facilitating consistent and comparable advancements across different research efforts in this domain.

\section{Conclusion}
In conclusion, this study presents a pioneering exploration of GNNs in the domain of supply chain management, addressing the current gap in research by connecting the inherently graph-like structure of supply chains with GNN methodologies. By formulating supply chains as graphs and leveraging GNNs, we have demonstrated significant performance improvements across multiple supply chain tasks, including regression, classification, detection, and anomaly detection. We introduce a new, multi-perspective benchmark dataset from a leading FMCG company in Bangladesh, which serves as a crucial resource for further research and analytics. This dataset provides a solid foundation for future studies in the application of GNNs to supply chain management. Our benchmarking demonstrates that GNN-based approaches consistently outperform traditional statistical and machine learning methods by 10-30\% in regression and classification tasks, and by 15-40\% in anomaly detection, underscoring their potential to enhance supply chain efficiency and decision-making. Furthermore, we discuss the real-world implications of applying GNNs to supply chains, considering their societal impacts and the limitations of our current study. This work not only lays the groundwork for future research but also highlights the transformative potential of GNNs in solving complex supply chain problems, ultimately contributing to the advancement of supply chain analytics and modelling.

\section*{Conflict of Interest}
The authors declare that there are no conflicts of interest regarding the publication of this paper. All research procedures followed ethical guidelines, and the study was conducted with integrity and transparency. The authors have no financial, personal, or other relationships that could inappropriately influence or bias the content of this work.

\section*{Data Availability}
Data related to this work is publicly available at \textbf{\textit{\href{https://doi.org/10.5281/zenodo.13652826}{\texttt{DOI: 10.5281/zenodo.13652826}}}} under the \href{https://creativecommons.org/licenses/by/4.0/legalcode}{CC BY 4.0} Licence.

\section*{Author Contributions}
ATW and MSI conceived the core idea, developed the methodology and theoretical aspects, prepared and processed raw data, conducted formal analysis, wrote and edited the paper, and led the project. 
ATW was responsible for designing and performing the experiments, conducting all analyses, and preparing visualizations. 
MSI handled the background study, data collection, and provided project supervision. 
ARA contributed to formal analysis and data collection. 
MMB provided guidance on manuscript preparation, writing, and revisions.

\bibliographystyle{tfcad}
\bibliography{our_work}


\appendix

\end{document}